\documentclass{template/llncs}
\usepackage{multirow}
\usepackage{csquotes}
\usepackage{booktabs}
\usepackage{graphicx}
\graphicspath{{images/}}
\usepackage{mathptmx}
\usepackage{times} 
\usepackage{amsmath} 
\usepackage{amssymb} 
\usepackage{subfigure}
\usepackage{float}
\usepackage{booktabs}
\usepackage{mathalfa}
\usepackage{caption}

\usepackage[mathcal]{eucal}
\usepackage[scientific-notation=true]{siunitx}
\usepackage{color}
\usepackage[dvipsnames]{xcolor}

\title{Enhancing clinical MRI Perfusion maps with data-driven maps of complementary nature for lesion outcome prediction}

\author{Adriano Pinto \inst{1,2} \and S\'ergio Pereira \inst{1,2} \and Raphael Meier\inst{3} \and Victor Alves\inst{2} \and Roland Wiest\inst{3}  \and Carlos A. Silva\inst{1} \and Mauricio Reyes\inst{4}}
\institute{CMEMS-UMinho Research Unit, University of Minho, Guimar\~{a}es, Portugal \email{id6376@alunos.uminho.pt} \and Centro Algoritmi, University of Minho, Braga, Portugal \and Support Center for Advanced Neuroimaging - Institute for Diagnostic and Interventional Neuroradiology, University Hospital Inselspital and University of Bern, Switzerland \and Institute for Surgical Technology and Biomechanics, University of Bern, Switzerland}

\begin{document}
\maketitle
\begin{abstract}

Stroke is the second most common cause of death in developed countries, where rapid clinical intervention can have a major impact on a patient's life. To perform the revascularization procedure, the decision making of physicians considers its risks and benefits based on multi-modal MRI and clinical experience. Therefore, automatic prediction of the ischemic stroke lesion outcome has the potential to assist the physician towards a better stroke assessment and information about tissue outcome. Typically, automatic methods consider the information of the standard kinetic models of diffusion and perfusion MRI (e.g. Tmax, TTP, MTT, rCBF, rCBV) to perform lesion outcome prediction. In this work, we propose a deep learning method to fuse this information with an automated data selection of the raw 4D PWI image information, followed by a data-driven deep-learning modeling of the underlying blood flow hemodynamics. We demonstrate the ability of the proposed approach to improve prediction of tissue at risk before therapy, as compared to only using the standard clinical perfusion maps, hence suggesting on the potential benefits of the proposed data-driven raw perfusion data modelling approach.

\end{abstract}

\section{Introduction}
	
	Stroke ranks second as leading cause of deaths worldwide, with ischemic stroke being the most common type. Ischemic stroke arises from a sudden occlusion of a cerebral artery. Diagnosis and treatment begins with the acquisition of multi-modal MRI or CT images, followed by appropriate medical intervention. While several clinical trials have proven the efficacy of mechanical thrombectomy, the treating physician must carefully evaluate the associated risks and benefits: the volume of ill-perfused tissue potentially salvageable, versus the risk of causing haemorrhage or other complications \cite{world2014global,wardlaw2010neuroimaging}. Hence, predicting the outcome of a stroke lesion (i.e. lesion status at three-month follow-up), and thereby evaluating the effect of a successful or unsuccessful mechanical thrombectomy, has a great potential to guide the decision of the physician. 
    
	MRI perfusion and diffusion sequences have been gaining importance in the characterization of ischemic stroke, which provides important information for the revascularization therapy \cite{barber1999identification,song2017temporal}. Based on MRI sequences, some machine learning approaches have been recently proposed for ischemic stroke lesion prediction. The majority is based on multivariate linear regression models \cite{scalzo2012,kemmling2015}, or through more advanced models such as decision trees \cite{mckinley} and CNN-based deep learning architectures \cite{maier2017isles}. However, up to our knowledge none of the approaches takes into consideration the temporal Perfusion Weighted Imaging data (4D PWI) for stroke lesion prediction. We hypothesize that a data-driven approach to model the raw perfusion imaging data can unveil information complementing the standard clinical perfusion maps derived using kinetic analysis.
    
   	In this paper, we propose a novel end-to-end deep learning multi-data branched network that incorporates information from the 4D PWI alongside the standard clinical perfusion and diffusion maps. From the time-stamp acquisitions of the 4D PWI, that characterize the bolus passage, we aim to learn brain blood flow hemodynamics (principal and collateral) to characterize tissue at risk of infarction (penumbra), and the unsalvageable tissue (ischemic core). Since standard perfusion and diffusion maps are generated from kinetic models followed by thresholding (based on clinical knowledge and experience), we hypothesize that there might be loss of relevant information. Hence, we aim to enhance the clinical MRI diffusion and perfusion maps with data-driven maps for stroke lesion outcome prediction. The proposed architecture was evaluated using the publicly-available ISLES 2017 dataset.
    
   	 The remainder of the paper is organized as follows: Section \ref{sec:methods} describes the methods of the proposed approach. Section \ref{sec:setup} addresses the setup for stroke lesion outcome prediction. Section \ref{sec:results} contains the obtained results and its discussion. Finally, Section \ref{sec:conclusions} contains the conclusions.

\section{Methods}
\label{sec:methods}
    
    \subsubsection{4D PWI} 
\begin{figure}[h]
		\centering
    	\includegraphics[width=0.95\textwidth]{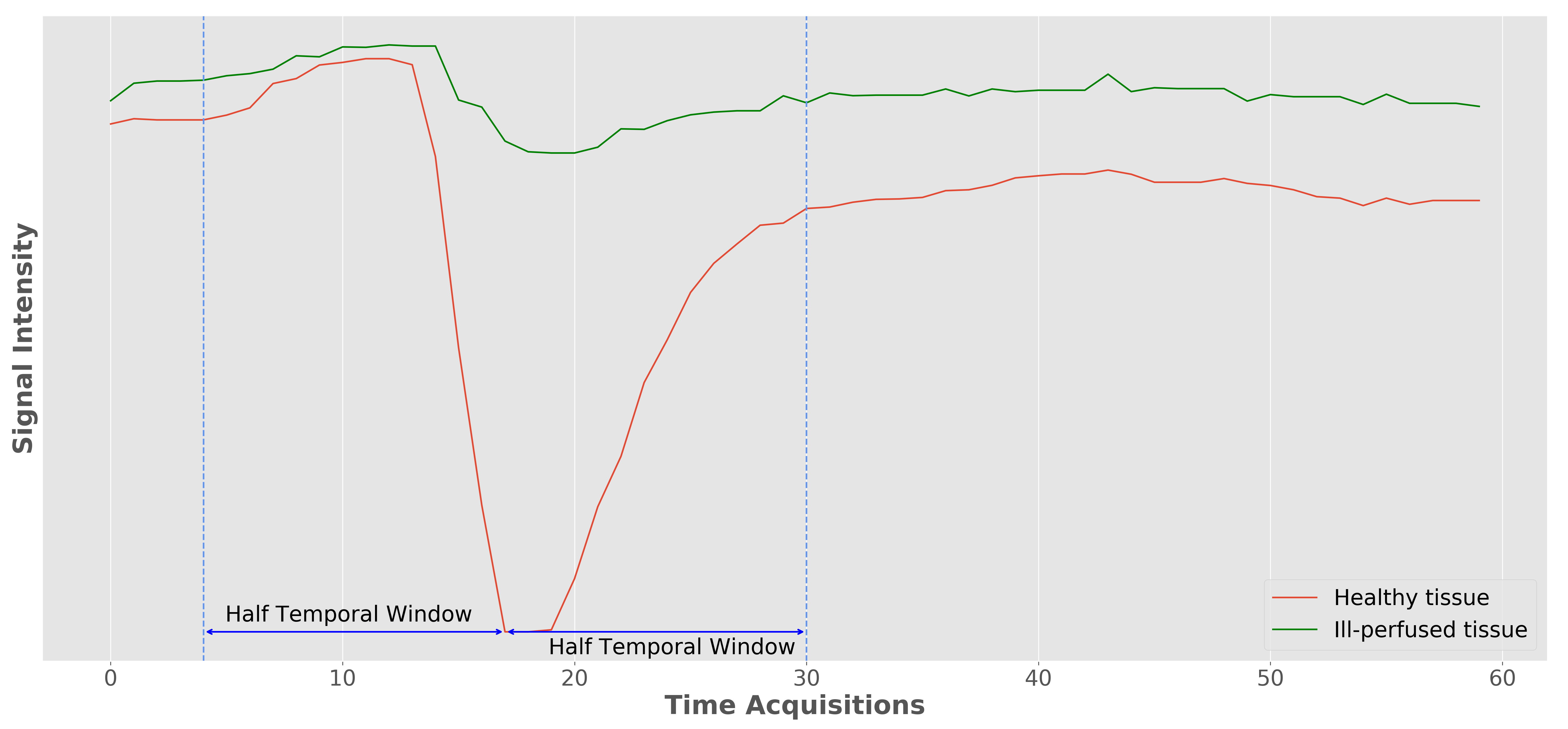}
		\caption{General 4D PWI signal intensity over time acquisitions in a patient with an ischemic stroke lesion. The dashed line corresponds to the temporal slicing performed automatically. The system automatically detects and selects a temporal window of interest, modeling temporal acquisitions characterizing blood dynamics of interest.}
        \label{fig:4DPWI_signal}
	\end{figure}

At the arrival of the contrast agent to the brain, during the acquisition of the 4D PWI, the healthy tissue will present a drop in the signal intensity value, which then increases as contrast agent starts diluting throughout the system. However, in the presence of ill-perfused tissue the signal intensity values barely changes, since there is no propagation of the contrast agent to the damaged tissue \cite{song2017temporal}. Figure \ref{fig:4DPWI_signal} depicts such signal intensity behaviour in a patient with ischemic stroke. The perfusion blood flow dynamic, captured by the temporal slices of the 4D PWI, is responsible for the generation of the 3D MRI perfusion maps, through the application of kinetic models, deconvolutions in the time space, and clinical thresholding. Therefore, rCBF, rCBV, MTT, TTP, and Tmax perfusion maps can be viewed as surrogate parametric summaries of the raw 4D PWI, encompassing specific blood flow dynamics. From this knowledge emerged the intention to evaluate the encoding of the blood flow hemodynamics directly from 4D PWI, considering altogether complementary information over the diffusion and perfusion maps. Our approach was based on the peak concentration of contrasting agent, which is of extreme importance, since it characterizes the point where the differences of perfusion between healthy tissue and the ill-perfused tissue are higher, allowing a better detection of the penumbra \cite{hosseini2017role}. We developed an automatic approach to detect the time slice where the concentration of contrast agent is higher, which corresponds to a lower signal intensity. We detect automatically the peak of concentration using k-means on the mean signal intensity and standard deviation. The peak is used to define a temporal window to retrieve specific temporal acquisitions regarding the blood dynamics needed to estimate the tissue at risk of infarction. Besides reducing the total number of temporal slices, we also enforce the same spatial-temporal space across patients. Aligning patient data across the peak concentration of contrasting agent yields a common time interval for the retrieval of information. A fixed temporal window size of 26 slices was used, based on the sampling rate of the MRI acquisition.

		\subsubsection{Baseline Architecture.} For stroke outcome lesion prediction, we based our network on the U-Net \cite{ronneberger2015u}, which has proved to be competitive in many biomedical image segmentation applications. The output of the U-Net is fed to a bi-dimensional GRU layer \cite{cho2014properties} that processes the information in four directions (superior-inferior, inferior-superior, anterior-posterior and posterior-anterior), to enforce a greater spatial context in stroke lesion outcome prediction. The baseline architecture only considers standard diffusion and perfusion MRI maps (as employed by the state of the art approaches).
    
	\subsubsection{Multi-Data Branched Network.} 
\begin{figure}[h]
		\centering
    	\includegraphics[width=\textwidth]{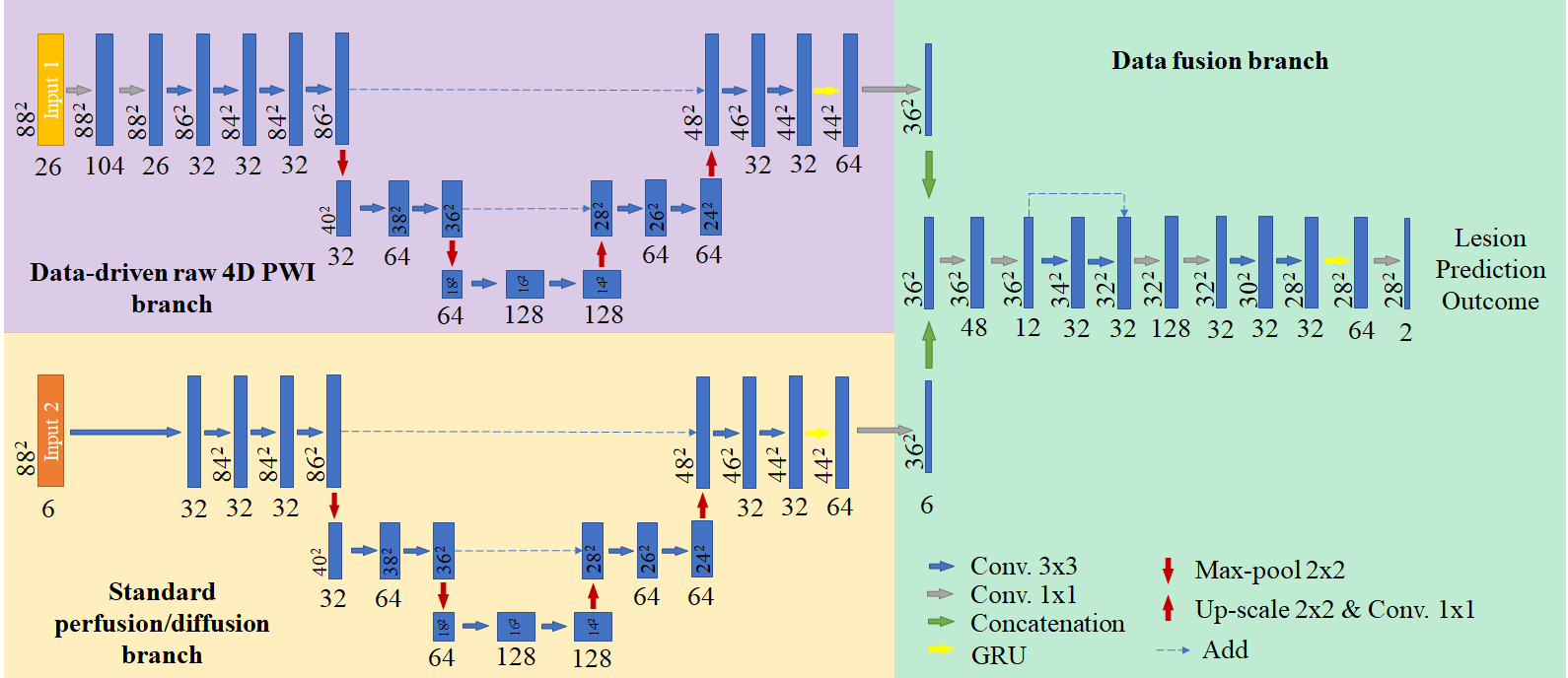}
		\caption{Overview of the proposed architecture for stroke lesion outcome prediction. Blue feature maps result from bi-dimensional convolutions.}
        \label{fig:nnet}
	\end{figure}

To merge information from the standard perfusion and diffusion maps and the data-driven 4D PWI data, the proposed architecture fuses two U-Nets as shown in Figure \ref{fig:nnet}. The top branch U-Net models the raw 4D PWI information, where the temporal information is coded as input channels. The first two layers consist of a feature expansion and feature reduction by 4. Recombining the feature maps allows complex interactions between temporal slices within the temporal window \cite{xie2017aggregated}. Since each branch is able to learn different specific features, we then merge the output of each branch, which is fed into a smaller architecture in order to take advantage from complementary information for stroke lesion outcome prediction. The final portion of the network encompasses also a bi-dimensional GRU layer present in both branches of the network.

\section{Setup}
\label{sec:setup}

	Our proposal was validated on the publicly available ISLES 2017 database, with a total of 75 cases divided into two datasets: Training dataset (n=43) and Challenge dataset (n=32). Each case contains a raw 4D PWI, five 3D MRI perfusion maps (rCBF, rCBV, MTT, TTP, Tmax), one 3D MRI diffusion map (ADC), and the final lesion outcome, which was manually segmented by a clinician on a 90-day follow-up (only available for the training dataset). All MRI maps are already co-registered and skull-stripped \cite{maier2017isles}. 
    
    Since MRI acquisitions are from different centers, all maps were resized to the same volume space: $256 \times 256 \times 32$.  $T_{max}$ was clipped to $\left[ 0, 20s \right]$, and the ADC was clipped to be within the range $\left[ 0, 2600 \right] \times 10^{-6} mm^{2}/s$, as values beyond these ranges are known to be biologically meaningless \cite{mckinley}. Afterwards, a linear scaling was performed between $\left[ 0, 255 \right]$. Bias field correction was performed to the 4D PWI using the N4ITK method \cite{tustison2010n4itk} before the resizing and scaling steps.

	The training dataset was divided into $36$ cases for training and $7$ cases for validation. For each case 550 patches of size $88 \times 88$ were randomly extracted. The network was trained with ADAM optimizer (learning rate$=1e-5$), with a mini-batch of size $4$, using as loss function the soft-dice loss \cite{milletari2016v}.  The sum is performed for the $N$ voxels of the patch both in the binary prediction $p_i \in P$ and the ground truth $g_i \in G$. The gradient of the Dice score for the $j-th$ voxel of prediction, was calculated as in Equation \ref{ref:soft_dice_loss}.
    
    \begin{equation}
		 	\frac{\delta Dice }{\delta p_j} = \frac{g_j (\sum_{i}^{N} p_{i}^{2} + \sum_{i}^{N} g_{i}^{2}) - 2 p_j \sum_{i}^{N} p_i g_i}
		 	{(\sum_{i}^{N} p_{i}^{2} + \sum_{i}^{N} g_{i}^{2})^2}
		\label{ref:soft_dice_loss}
	\end{equation}
    
    The tests were conducted using Keras with Theano, on a Nvidia GeForce GTX-1070 8GB, where each prediction took around 30s. 
	
	We compare the performance of our proposal with three different studies: Standard Branch, Data-Driven Branch, and Multi-Data Single Branch. The Standard Branch architecture considers diffusion and perfusion maps. The Data-Driven Branch studies the 4D PWI. In the Multi-Data Single Branch we combined the inputs from both branches into a single network. In Table \ref{tab:Results} we report results on ISLES 2017 test dataset, which enables us to compare with state of the art methods.

\begin{table}[htp]
  	\caption{Methods from ISLES 2017 testing dataset. Each metric contains the mean $\pm$ standard deviation.}

      \tiny
      \centering
      \resizebox{\textwidth}{!}{
      \begin{tabular}{ccccccc}
      \specialrule{.2em}{.1em}{.1em}
      && \multicolumn{1}{c}{Dice} 
      & \multicolumn{1}{c}{Hausdorff Distance} 
      & \multicolumn{1}{c}{ASSD}
      & \multicolumn{1}{c}{Precision}
      & \multicolumn{1}{c}{Recall}
      \\ \cmidrule(l){3-7}
       \multirow{16}{*}{\rotatebox[origin=c]{90}{Challenge}}  
       & Mok et al.* & 0.32 $\pm$ 0.23 & 40.74 $\pm$ 27.23 & 8.97 $\pm$ 9.52 & 0.34 $\pm$ 0.27 & 0.39 $\pm$ 0.27 \\
       & Kwon et al.* & 0.31 $\pm$ 0.23 & 45.26 $\pm$ 21.04 & 7.91 $\pm$ 7.31 & 0.36 $\pm$ 0.27 & 0.45 $\pm$ 0.30 \\
       & Bertels et al.* & 0.30 $\pm$ 0.21 & 33.85 $\pm$ 16.82 & 6.81 $\pm$ 7.18 & 0.34 $\pm$ 0.26 & 0.51 $\pm$ 0.32 \\
       & Monteiro et al.* & 0.30 $\pm$ 0.22 & 46.60 $\pm$ 17.50 & 6.31 $\pm$ 4.05 & 0.34 $\pm$ 0.27 & 0.51 $\pm$ 0.30 \\
       & Lucas et al.* & 0.29 $\pm$ 0.21 & 33.85 $\pm$ 16.82 & 6.81 $\pm$ 7.18 & 0.34 $\pm$ 0.26 & 0.51 $\pm$ 0.32 \\
       & Choi et al.* & 0.28 $\pm$ 0.22 & 43.89 $\pm$ 20.70 & 8.88 $\pm$ 8.19 & 0.36 $\pm$ 0.31 & 0.41 $\pm$ 0.31 \\
       
       & Robben et al.* & 0.27 $\pm$ 0.22 & 37.84 $\pm$ 17.75 & 6.72 $\pm$ 4.10 & 0.44 $\pm$ 0.32 & 0.39 $\pm$ 0.31 \\
       & Pisov et al.* & 0.27 $\pm$ 0.20 & 49.24 $\pm$ 32.15 & 9.49 $\pm$ 10.56 & 0.31 $\pm$ 0.27 & 0.39 $\pm$ 029 \\
       & Niu et al.* & 0.26 $\pm$ 0.20 & 48.88 $\pm$ 11.20 & 6.26 $\pm$ 3.02 & 0.28 $\pm$ 0.25 & 0.56 $\pm$ 0.26 \\
       & Sedlar et al.* & 0.20 $\pm$ 0.19 & 58.30 $\pm$ 20.02 & 11.19 $\pm$ 9.10 & 0.23 $\pm$ 0.24 & 0.40 $\pm$ 0.29 \\
       & Rivera et al.* & 0.19 $\pm$ 0.16 & 63.58 $\pm$ 18.58 & 11.13 $\pm$ 7.89 & 0.27 $\pm$ 0.25 & 0.21 $\pm$ 0.17 \\
       & Islam et al.* & 0.19 $\pm$ 0.18 & 64.15 $\pm$ 28.51 & 14.17 $\pm$ 15.80 & 0.29 $\pm$ 0.28 & 0.25 $\pm$ 0.25 \\
       & Chengwei et al.* & 0.18 $\pm$ 0.17 & 65.95 $\pm$ 25.94 & 9.22 $\pm$ 6.99 & 0.37 $\pm$ 0.30 & 0.21 $\pm$ 0.23 \\
       & Yoon et al.* & 0.17 $\pm$ 0.16 & 45.23 $\pm$ 19.14 & 12.43 $\pm$ 11.01 & 0.23 $\pm$ 0.27 & 0.36 $\pm$ 0.32 \\

       \cmidrule(lr){2-7}
       & Standard Branch & 0.20 $\pm$ 0.19 & 70.04 $\pm$ 20.37 & 11.66 $\pm$ 7.40 & 0.16 $\pm$ 0.20 & 0.61 $\pm$ 0.28 \\
       & Data-Driven Branch & 0.20 $\pm$ 0.18 & 54.59 $\pm$ 18.69 & 9.95 $\pm$ 5.79 & 0.18 $\pm$ 0.21 & 0.61 $\pm$ 0.27 \\
       & Multi-Data Single Branch & 0.26 $\pm$ 0.21 & 46.35 $\pm$ 17.59 & 8.37 $\pm$ 6.43 & 0.21 $\pm$ 0.20 & 0.61 $\pm$ 0.28 \\
       & Multi-Data Branched & 0.29 $\pm$ 0.21 & 41.58 $\pm$ 22.04 & 7.69 $\pm$ 5.71 & 0.23 $\pm$ 0.21 & 0.66 $\pm$ 0.29 \\				
      \bottomrule
      \noalign{\vskip 2pt}  
      \multicolumn{6}{l}{$^{*}$ Static results from \cite{Site}.} \\
      \end{tabular}}
  	\label{tab:Results}
\end{table}
\section{Results and Discussion}
\label{sec:results}
   	
    Table \ref{tab:Results} contains the results obtained on the ISLES 2017 testing dataset. The Standard Branch and the Data-Driven Branch achieved the same Dice score, but with differences in the distance metrics. The Data-Driven Branch was capable of predicting the lesion outcome with higher robustness, since the Hausdorff and ASSD are lower when compared to the Standard Branch. Nevertheless, both approaches are not capable of reaching state of the art performance. However, when we fuse the information of both models, as proposed in this paper, we observe an improvement on the average Dice score, but also on Hausdorff and ASSD. Being so, both models provide distinct information of value for stroke lesion outcome prediction. Additionally, we also study the performance of a single model with all the inputs combined, referred to as the Multi-Data Single Branch architecture. Such approach reached higher Dice score than the two branches separately, and with lower distance metrics. However, it was not capable of reaching the same performance of our proposal. Therefore, having 2 U-Nets for different input data has benefits on modularity and specificity on how the information is modelled. Using separate deep learning models to directly learn intrinsic biological phenomena allowed a higher robustness and accuracy in lesion outcome location and delineation, which is sustained by the lower distance metric values and higher Dice score.
    
\begin{figure}[h]
		\centering
    	\includegraphics[width=1.0\textwidth, keepaspectratio]{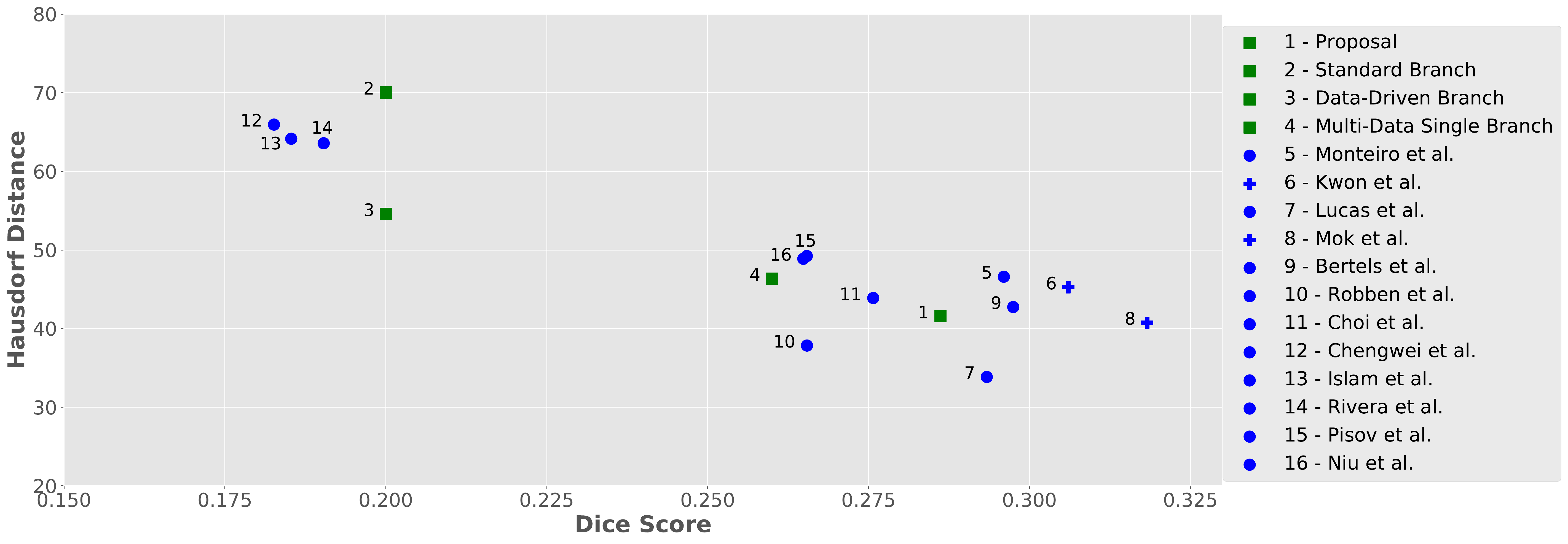}
		\caption{Hausdorff Distance versus Dice score from methods of ISLES 2017 in the testing set. Note that closer to the x axis and further away from the origin is better. Ensemble methods are marked with a cross.}
        \label{fig:HD_vs_Dice}
	\end{figure}

        In addition, we compare our proposal with other approaches. However, such comparison needs to consider the fact that top rank approaches use multiple models (i.e. ensembling). Our proposal was able to achieve a Dice score among the top 5 ranking methods, being within the same Hausdorff range of those methods, with just a single network. From this comparison we highlight the low distance metric values obtained, and also a Dice score in the same level of the 4th ranked method. Figure \ref{fig:HD_vs_Dice} shows the average Dice score and respective Hausdorff for each method.   
 
\begin{figure}[h]
		\centering
    	\includegraphics[width=\textwidth]{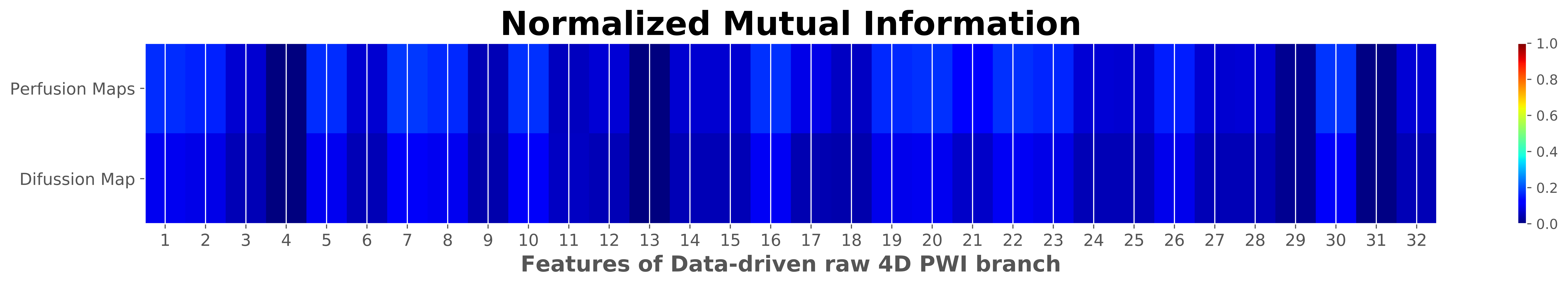}
		\caption{Normalized Mutual Information between the standard perfusion/diffusion maps and the feature maps from the data-driven branch, in training case 36. Values closer to 0 mean low mutual information and closer to 1 represent a high association.}
        \label{fig:MI}
	\end{figure}

  	From this analysis we emphasize the benefits of the proposed approach to extract and model information that might not be fully characterized by the standard perfusion and diffusion maps. To assess the complementarity of the data-driven perfusion maps, we computed the normalized mutual information between all the standard perfusion maps and each of the learned feature maps from the data-driven raw 4D PWI branch. As shown in Figure \ref{fig:MI}, low association values (less than 0.2) were obtained for all the extracted feature maps, meaning that both branches introduce new and complementary features.
    
\begin{figure}[!htb]
\centering
\subfigure[]{\label{subfig:training_33}\includegraphics[width=0.75\textwidth, keepaspectratio]{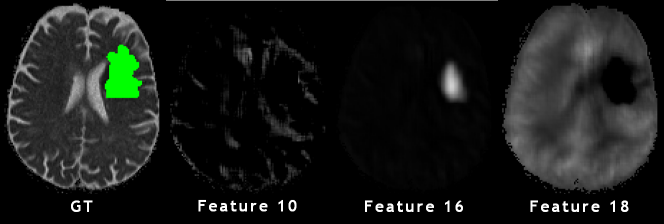}}
\subfigure[]{\label{subfig:training_36}\includegraphics[width=0.75\textwidth, keepaspectratio]{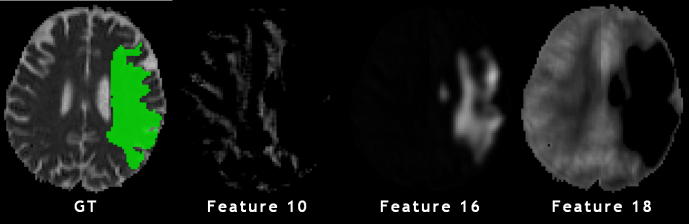}}
\caption{Example of extracted features for training case 33 (a) and case 36 (b), alongside the corresponding ground-truth overlapped over the ADC map.}
\label{fig:feature_maps}
\end{figure}

    \newpage
    Figure \ref{fig:feature_maps} shows the extracted features from the data-driven raw 4D PWI branch. Visually, feature 10 can reflect some descriptions of collateral blood flow, where features 16 and 18 focus on the surrounding lesion area itself in a complementary way. This analysis is particularly important since it can provide complementary information in the decision making process performed by the clinician, providing a better prediction of potentially salvageable tissue.

\section{Conclusions and Future Work}
\label{sec:conclusions}
        
     Parametric perfusion maps can be affected by intrinsic patient physiology \cite{song2017temporal}. To cope with this effect mathematical models are applied to standardize the behaviour of the contrast bolus passing. Nonetheless, it cannot be independent from patient specific blood flow hemodynamics, which can highly affect the perfusion parametric maps by adding a wide variability in the penumbra delineation \cite{song2017temporal}. Therefore, in this work, we propose a deep learning architecture, that can process the information from raw 4D PWI data and generate complementary information to the perfusion parametric sequences, which as shown here can increase stroke lesion outcome prediction.
        
    In the future, we intend to perform an interpretability analysis of the data-driven perfusion maps as well as an analysis of the learning patterns of the architecture to ensure the correctness of the predictions with respect to the data being used to drive the lesion outcome predictions.
    
\subsubsection{Acknowledgments} Adriano Pinto was supported by a scholarship from the Funda\c{c}\~ao para a Ci\^encia e Tecnologia (FCT), Portugal (scholarship number  PD/BD/113968/2015). This work has been supported by COMPETE: POCI-01-0145-FEDER-007043 and FCT – Funda\c{c}\~ao para a Ci\^encia e Tecnologia within the Project Scope: UID/CEC/00319/2013.

\bibliographystyle{template/splncs03} 
\bibliography{references}

\end{document}